\def\year{2022}\relax
\documentclass[letterpaper]{article} 
\usepackage{aaai22}  
\usepackage{times}  
\usepackage{helvet}  
\usepackage{courier}  
\usepackage[hyphens]{url}  
\usepackage{graphicx} 
\usepackage{natbib}  
\usepackage{caption} 
\DeclareCaptionStyle{ruled}{labelfont=normalfont,labelsep=colon,strut=off} 
\usepackage{algorithm}
\usepackage{algorithmic}
\usepackage{multirow}

\usepackage{newfloat}
\usepackage{listings}
\usepackage{todonotes}
\usepackage{subcaption}
\newcommand\Tstrut{\rule{0pt}{2.6ex}}         
\newcommand\Bstrut{\rule[-0.9ex]{0pt}{0pt}}   
\floatstyle{ruled}
\newfloat{listing}{tb}{lst}{}
\floatname{listing}{Listing}

\title{AnnoBERT: Effectively Representing Multiple Annotators' Label Choices to Improve Hate Speech Detection}
\author{
    Wenjie Yin\equalcontrib,\textsuperscript{\rm 1}
    Vibhor Agarwal\equalcontrib,\textsuperscript{\rm 2}
    Aiqi Jiang\equalcontrib,\textsuperscript{\rm 1}
    Arkaitz Zubiaga,\textsuperscript{\rm 1}
    Nishanth Sastry\textsuperscript{\rm 2}
}
\affiliations{
    \textsuperscript{\rm 1}Queen Mary University of London, UK\\
    \textsuperscript{\rm 2}University of Surrey, UK

}

\usepackage{bibentry}

\begin{document}

\maketitle

\begin{abstract}
Supervised approaches generally rely on majority-based labels. However, it is hard to achieve high agreement among annotators in subjective tasks such as hate speech detection. Existing neural network models principally regard labels as categorical variables, while ignoring the semantic information in diverse label texts. In this paper, we propose AnnoBERT, a first-of-its-kind architecture integrating annotator characteristics and label text with a transformer-based model to detect hate speech, with unique representations based on each annotator's characteristics via Collaborative Topic Regression (CTR) and integrate label text to enrich textual representations. During training, the model associates annotators with their label choices given a piece of text; during evaluation, when label information is not available, the model predicts the aggregated label given by the participating annotators by utilising the learnt association. The proposed approach displayed an advantage in detecting hate speech, especially in the minority class and edge cases with annotator disagreement. Improvement in the overall performance is the largest when the dataset is more label-imbalanced, suggesting its practical value in identifying real-world hate speech, as the volume of hate speech in-the-wild is extremely small on social media, when compared with normal (non-hate) speech. Through ablation studies, we show the relative contributions of annotator embeddings and label text to the model performance, and tested a range of alternative annotator embeddings and label text combinations. 
\end{abstract}

\section{Introduction}

 Hate speech is defined as speech that directly attacks or promotes violence towards an individual or a group based on actual or perceived aspects of personal characteristics, such as ethnicity, gender and religion \citep{waseem2016hateful,davidson2017automated,fortuna2018survey,yin2021towards}.
Hate speech has become a pressing problems for all major social media platforms and its prevalence has brought about the need to come up with automated solutions \citep{fortuna2018survey,jurgens2019just}. 

Most studies tackle the problem of hate speech detection from an algorithmic perspective and state-of-the-art approaches rely more on supervised learning \citep{badjatiya2017deep,schmidt2017survey,fortuna2018survey,arango2020hate,jiang2021cross}, where the agreement level can impact the quality of data resources, potentially leading to model performance issues.
Given its subjective nature, it is difficult to achieve high inter-rater agreement when labelling hate speech \citep{kocon2021offensive}. Online hate speech often varies widely across groups, communities and countries \citep{kocon2021offensive}. Effective identification of hate speech strongly depends on the annotator's world view \citep{akhtar2021whose}, leading to lower inter-rater agreement than more objective classification tasks \citep{davani2021dealing}. Differences among annotators with different demographic and cultural backgrounds can in turn be useful, providing deeper insights into the content \citep{waseem2016you,davani2021dealing}.
However, these differences are often overlooked while aggregating all annotations into a majority label \citep{beelen2017detecting}.

Taking into account annotator perspectives is an emerging direction in subjective tasks, and consequently in hate speech or offensive language detection. 
Existing works usually first separate annotators into groups according to demographic features \citep{binns2017like,wich2020investigating} or personal views \citep{akhtar2020modeling,akhtar2021whose} and then train classifiers on each group, or use annotator metadata as part of model input \citep{kocon2021offensive}.

Despite consistent attempts to incorporate annotators' perspectives, most related studies \citep{akhtar2021whose,kocon2021offensive} overlook another attribute inside annotated datasets - explicit information from labels. They only implicitly learn the underlying content of the labels from training data by treating the labels as categorical variables, ignoring the latent linguistic knowledge of the words used for the labels. Since online texts from social media might not be long enough to convey comprehensive information \citep{guo2013linking}, label information could be advantageous to improve the performance of neural text classification models by constructing the text-sequence representations, enriching the textual representations and reducing the ambiguity from input texts \citep{wang2018joint,pappas2019gile}. \citet{wang2018joint} firstly embed text and label information into the joint space and effectively enhance the text classification performance. Recent studies also demonstrate the capacity of capturing complex label relationships to improve semantic representations of input text in text classification \citep{pappas2019gile,halder2020task,luo_dont_2021}. More recent works effectively capture the conversational context in online conversations for text classification tasks such as polarity prediction~\citep{agarwal2022graphnli,young2022modelling} and hate speech detection~\citep{agarwal2022graph}.

Given the potential impact of annotator perspectives and labelling history, this paper presents a novel architecture, AnnoBERT, a first-of-its-kind model that effectively incorporates label choices from different annotators into the hate speech detection task without the need to train separate classifiers or use annotator metadata.
In order to obtain text representations containing more comprehensive text semantics from annotators' behaviours, AnnoBERT models annotators' preferences from their annotation history and associates such preferences with the specific hate speech annotation tasks through label texts. Our AnnoBERT architecture yields performance gains over the base transformer model and other baselines for the hate speech detection task on two commonly-used datasets and demonstrates its detection superiority on class-imbalanced datasets. In addition, our ablation study results indicate that, compared to other alternative embeddings, the CTR annotator embedding shows its benefit, and label semantics with high-contrast text also boosts the performance of AnnoBERT for identifying online hate. 

Our main contributions are the following:

\begin{enumerate}
    \item We introduce the first integrated model\footnote{Our code will be made available upon publication.} to associate multiple annotator perspectives to text and label representations for hate speech detection;
    \item We utilise an intuitive and explainable model to construct unique behaviour representations for each annotator based on the previous annotations;
    \item We demonstrate the advantage of AnnoBERT compared to all baselines and its benefit for the minority class in hate speech data;
    \item We show the relative contribution to the overall performance of modelling annotator preferences and label texts through ablative experiments.
\end{enumerate}

\section{Background and Related Work}

\subsection{Multiple Annotator Perspectives}


To perform the annotation aggregation discussed above, several well-known methods, such as probabilistic models of annotation \citep{plank2014adapting,paun2018probabilistic} and neural-based aggregation methods \citep{yin2017aggregating,rodrigues2018deep,li2020neural}, have been applied to address this problem. 

Due to imprecise or vague annotation schemes, however, disagreements among annotators have universally appeared \citep{paun2022statistical}, especially in highly subjective annotation projects, such as sentiment analysis \citep{kenyon-dean2018sentiment}, abusive language detection \citep{akhtar2020modeling} and hate speech detection \citep{agarwal2022graph}. Then, instead of only aggregating annotations based on majority vote (hard labels), some studies start to directly learn from a distribution of annotation labels without aggregation (soft labels) \citep{peterson2019human,rodrigues2018deep,chou2019every,Uma2020case,fornaciari2021beyond,Chu2021learning}, which can be modelled jointly with hard labels \citep{chou2019every}, incorporated into the loss function \citep{plank2014learning,plank2016multilingual,Uma2020case}, learnt to generate representations of annotators \citep{rodrigues2018deep}, integrated as an auxiliary task \citep{fornaciari2021beyond}, or learnt with the context \citep{Uma2020case}. \citet{Chu2021learning} propose a similar end-to-end approach, with two types of annotation noise layers, modelling annotator confusion in terms of expertise and instance difficulty.


In the increasing body of research exploring annotator agreement in hate speech, some have focused on exploring how to aggregate annotations based on demographic characteristics \citep{binns2017like,al2020identifying,wich2020investigating} or annotation history \citep{akhtar2020modeling,akhtar2021whose}, resulting in classifiers tailored to each group. 
Other studies have tended to directly adopt the distribution of annotations into a multi-task \citep{davani2021dealing} or a multi-modal \citep{kocon2021offensive} architecture.

Demographic data is usually unavailable and could be incomplete or imprecise even when available, as a result of a crowdsourcing task where identity isn't verified. Dividing annotators into groups also has its shortcoming --- the number of groups can be arbitrary; training a classifier for each group is inefficient. Hence, we produce annotator representations based on their annotation history, which, without grouping, are directly exploited as part of the input for the hate speech detection model.

\subsection{Label Information}

A growing number of works related to the text classification task has recently realised the value of label information in improving textual representations \citep{wang2018joint,pappas2019gile,halder2020task,luo_dont_2021}. Work on modelling label information can be divided into two types.

On one hand, labels can be modelled with label embeddings, as another input independent of text embeddings. An attentive architecture with joint label embeddings is used to fuse texts and labels into a space and learn label-attentive representations by assigning attention score to the texts with homologous labels \citep{wang2018joint}.
\citet{pappas2019gile} also learns joint input-label embeddings by encoding label descriptions and texts separately, and projecting them into a joint input-label multiplicative space for classification.

On the other hand, instead of being modelled separately, label information can be integrated with the input text. Prompt-based methods \citep{schick_2021_exploiting} encode label and task information in the form of questions as part of language model input.
\citet{luo_dont_2021} simplified this by using one single label word instead of manually constructing questions as model input, and \citet{halder2020task} also use label words and text pairs as the input to pre-trained language models to learn a task-aware representation of sentences.
We adapt \citet{luo_dont_2021}'s approach to our model.  

Annotation bias exists for certain sub-types of hate speech \citep{waseem2016you,davidson2019racial,davani2021dealing}, thus the task label (e.g. ``misogyny", ``hateful") carry significant information for the hate speech detection task. However, existing hate speech detection models primarily adopt a binary prediction task, mapping a fixed set of categorical task labels into k-hot vectors. To some extent, they disregard semantic information in the label text, as well as the influence of the behaviours of multiple annotators on the single label of each text.
Therefore, we propose an intuitively motivated hate speech detection model, the first model utilising task label information to enrich input text representations and annotator perspective representations in order to boost the detection performance.

\section{Methodology}
Here we describe our proposed model -- AnnoBERT.

\subsection{Collaborative Topic Regression for Annotator Representation}

The Collaborative Topic Regression (CTR) model \citep{wang_collaborative_2011} is a probabilistic model originally used in recommendation systems. It discovers a set of topics from the corpus and models user preference with regards to the topics with their ratings of items in the corpus, enabling prediction of user ratings of unseen items based on the topics present. In the fitted model, each element in a user latent vector represents the user's preference for a topic, 
despite not having any explicit information about the topic itself.

We chose to adapt CTR to hate speech data based on the analogy that hate speech was also generated by a set of topics, and that annotators are expected to have different levels of sensitivity to each topic. We fitted CTR models using an open-source implementation\footnote{https://github.com/dongwookim-ml/python-topic-model} on the training set of each dataset for the default setting of 100 expectation-maximisation iterations, and used the user latent vectors as annotator embeddings. The non-parallelised and iterative implementation required 2 minutes per iteration on the \textit{Guest} dataset and 11 minutes on the \textit{Dynabench} dataset respectively on a single CPU. We experimented with a range of dimensions (5 to 15, as 10 is the default setting) and aggregation strategies (mean-pooling, max-pooling, summation, and concatenation), in terms of macro F1 on the validation set. We found 10-D latent vectors and mean-pooling to be the best performing option with moderate advantage over other options (1.8 to 9 macro F1); these settings are used in the final AnnoBERT model for all downstream tasks.

\subsection{Integrating Label Text}
Besides the annotator embeddings, we also leverage the annotators' chosen label text to associate relevant annotators to their opinions on a particular instance. For each entry in the dataset, we create two separate instances: one containing the annotators who labelled it as hate and the other containing the annotators who labelled it as non-hate, unless all annotators agree. During training, we then append the annotated class label text used by the specific dataset at the end for each of these instances, separated by the \texttt{[SEP]} token. This class label is not appended in the development and test set. The intuition is that during training, the model associates annotators with their label choices given a piece of text; so that during evaluation, when label information is not available, the model predicts the aggregated label given by the participating annotators by utilising the association learnt during training. The default model uses ``misogynistic'' or ``hate'' for the positive class, domain-specific to the original datasets as described in the ``Datasets'' section, and the label ``not'' for the negative class, assuming that non-hateful instances across datasets are not qualitatively different. 

\subsection{AnnoBERT Architecture}
\begin{figure}[h]
  \centering
  \includegraphics[width=0.8\linewidth]{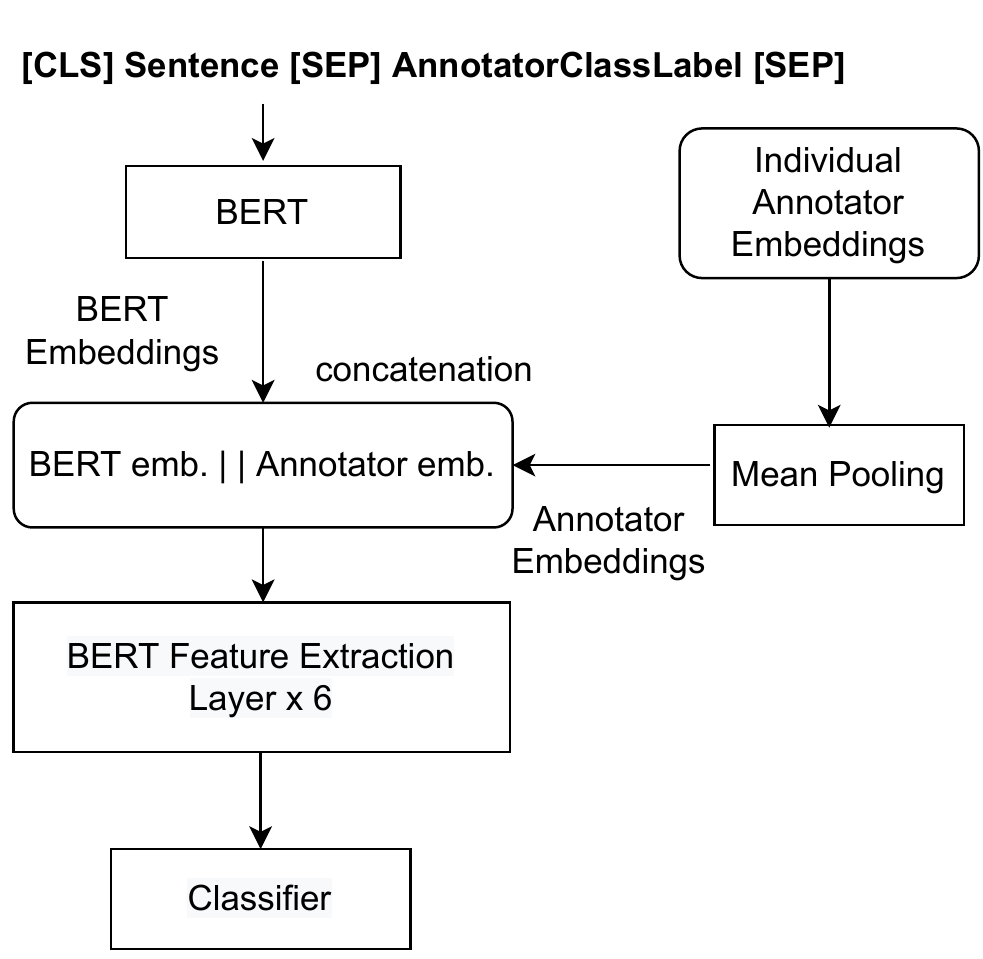}
  \caption{AnnoBERT Architecture. \texttt{||} = concatenation.}
  \label{fig:annobert}
\end{figure}

We input the annotators' class labels and annotator embeddings into the AnnoBERT model (see Figure~\ref{fig:annobert}). The sentence and the corresponding annotators' class label, separated by the \texttt{[SEP]} token, are input into a BERT encoder. The contexual embeddings are extracted from the final layer, and then concatenated with the aggregated annotator embeddings from those who annotated a particular instance with the same class label. These concatenated embeddings are then input into six BERT feature extraction layers, following \citet{zhang_ma-bert_2021} who found six feature extraction layers to be the best. Finally, the resultant embeddings are fed into the classification layer.
We used the most widely acknowledged Bidirectional Encoder Representations from Transformers (\texttt{BERT-base-uncased}) \citep{devlin2019bert} as the base model in our experiments, but the choice of the base model is completely flexible and can be applied on any other pretrained BERT-like model.

\section{Experiment Setup}

\subsection{Datasets}\label{sec:datasets}

\begin{table}[htb]
\centering
\begin{tabular}{llcc}
\hline
\multicolumn{2}{l}{} & \textbf{Guest} & \textbf{Dynabench}      \\
\hline
\textbf{Number of} & total & 6124      & 41094     \\
\textbf{Instances}   & non-hate & 5612     & \textbf{22132}    \\
             & hate   & \textbf{512}   & 18962  \\
\hline
\multicolumn{2}{l}{\textbf{Label}}                                                                   & \begin{tabular}[c]{@{}c@{}}misogynistic,  \Tstrut \\ nonmisogynistic\end{tabular} & \begin{tabular}[c]{@{}c@{}}hate, \\ nothate\end{tabular} \\
\hline
\end{tabular}%
\caption{Dataset statistics, with minority class bolded. }
\label{tab:datasets}
\end{table}

\begin{table}[htb]
\centering
\begin{tabular}{llcc}
\hline
\multicolumn{2}{l}{}                                                                                 & \textbf{Guest}                                                           & \textbf{Dynabench}       \\
\hline
\multicolumn{2}{l}{\textbf{Number of Annotators}}                                                    & 6*    & 20            \\
\hline
\multicolumn{2}{l}{\textbf{Annotators per Instance}}                                                 & 3    & 1              \\
\hline
\multirow{2}{*}{\textbf{Gender}}          & Female   & 4     & 13  \\
                          & Male     & 1    & 7    \\
\hline
\multirow{4}{*}{\textbf{Education}}  & HS       & 2      & 2   \\
  & UG       & 2     & 4     \\
  & PG       & 1  & 9   \\ 
  & PhD      & 0  & 5   \\
\hline
\multirow{2}{*}{\textbf{Nationality}}                                                     & British  & 5                                                                        & 12         \Tstrut         \\  
       & Other    & 0      & 8      \\
\hline
\end{tabular}%
\caption{Annotator statistics. Education background is listed, namely high school (HS), undergraduate (UG), postgraduate (PG), and research degree (PhD). *Five of the six annotators in \textit{Guest} give permission to share their basic information.}
\label{tab:annotator}
\end{table}

Due to the limited availability of annotators' meta-information and labels, in our work, we choose two contrasting datasets (see Table \ref{tab:datasets}), with the number of instances of each class and other features displayed.

\begin{itemize}
  \item \textit{Guest} \citep{guest_expert_2021}: Expert annotated, sourced from Reddit. Non-hate instances are the overwhelming majority. The domain and the official positive label text is ``misogynistic''. Each instance is annotated by 3 annotators on average. 6 unique annotators in total.

  \item \textit{Dynabench} \citep{vidgen_learning_2020}: Crowd generated and annotated, deliberate synthetic data designed to fool classifiers. No specific hate type, hence hate label is ``hate''. Slightly more data is labeled as hate speech than as non-hate. Each instance is only associated with one annotator who constructed the instance originally, but in some cases the class label could have been changed by an expert. This dataset has 20 unique annotators in total. 
\end{itemize}

These two datasets were, at the time when we started the experiments, the only datasets with sufficient information about the decisions of each annotator.
Table \ref{tab:annotator} summarises annotators' details in the two datasets. All annotators regularly use social media at least once per day, and more than half of them reported to have been targeted personally by online abuse before.
We use the official training and testing splits from the original dataset release, and further randomly sample 10\% from the training split to form the validation set. 

Text preprocessing was minimal: 
All usernames and urls were replaced with special tokens (``$<$user$>$", ``$<$url$>$"). Non-ascii letters were considered non-English and thus removed for being out of scope.  Additional whitespaces were removed.

\subsection{Baselines}

We compare AnnoBERT with three baseline models:

\begin{itemize}
  \item \textbf{BERT} \citep{devlin2019bert}:
Bidirectional Encoder Representations from Transformers (BERT) is a Transformer network \citep{vaswani2017attention} pre-trained with a language modelling objective and a vast amount of raw text. Having been acknowledged as a top performing model on many hate speech datasets, it was chosen as a representative of the BERT-like transformers. 

  \item \textbf{CrowdLayer} \citep{rodrigues2018deep}:
It is a neural network layer, which enables models to be trained end-to-end directly from the distribution of annotations. Such a layer is added on top of a BERT classifier to capture biases of multiple annotators and classification results can be obtained from the BERT classifier after the training stage.

  \item \textbf{LEAM} \citep{wang2018joint}:
Label-Embedding Attentive Model (LEAM) is an attention-based architecture with joint label embeddings, which embeds the text and label information into the same space to learn label-attentive text representations.
\end{itemize}

\subsection{Imbalanced sample (Dynabench)}

\textit{Guest} is much smaller and imbalanced than \textit{Dynabench}. This is a major but not the only difference between the two datasets. To isolate the effect of class imbalance on the relative performances, we randomly sampled from \textit{Dynabench} 6124 instances (512 hate, 5612 non-hate) to form a sample with the same class imbalance as \textit{Guest}.

\subsection{Disagreement subset (Guest)}

One might expect that taking into account annotator characteristics help clarify edge cases which annotators disagree on. To verify this hypothesis, we provide comparison of the original AnnoBERT model and the strongest baseline, BERT, on instances where disagreement between annotators are present in \textit{Guest}\footnote{The same experiment was not available for \textit{Dynabench}, in which each instance was only associated with one annotator.}, compared to the overall data. Possibly unsurprisingly, most instances with disagreement were predominantly deemed hateful; all instances without disagreement were consistently labelled as pertaining to the negative class. Thus, we did not include the subset without disagreement as the metrics would be obsolete when one class is absent. 
Note that the subset distinction only happens at the evaluation stage, using the models trained on the entirety of data. We present results on the entire dataset as well as on the disagreement subset only, for more detailed analysis of edge instances vs all instances.

Their sizes and the ratios of the positive class of subsets are shown in Table \ref{tab:subset_stats}.

\begin{table}
\centering
\begin{tabular}{l|cc} 
\hline
Subset                 & \multicolumn{1}{c}{Total} & \multicolumn{1}{c}{Hate ratio (\%)} \Tstrut\Bstrut \\ 
\hline
Guest, disagreement    & 170                             & 60.00                                       \Tstrut  \\
Guest, agreement & 1,061 & 0.00 \\
Guest, overall         & 1,231                            & 8.29                                       \Bstrut  \\
\hline
\end{tabular}
\caption{Subsets investigated with total instance counts and hate class ratio in the testing data. }
\label{tab:subset_stats}
\end{table}

\subsection{Model ablation: annotator embeddings}

We use AnnoBERT variants with two other types of annotator embeddings to test the effectiveness of the CTR-generated annotator embeddings. These model variants have the same architecture as the original AnnoBERT model (Figure \ref{fig:annobert}), with only the individual annotator embeddings replaced.
\begin{itemize}
     \item \textbf{History}. Instead of the CTR annotator embeddings, we insert raw annotation history as annotator vectors. In line with \citet{basile_its_2021}, each annotator vector is made of their annotations across all instances in the training data: -1 for non-hate, 1 for hate, 0 when the annotator didn't label that instance. The dimensionality of the vector is then linearly reduced to the same as BERT hidden size.
     Models with the embedding layer weights frozen and unfrozen were experimented.
    \item \textbf{Learnt}. To test whether annotation preference offers sufficiently more useful information than annotator identifiers, we use a model which randomly initialises an embedding layer for each annotator and learns the annotator embeddings end-to-end. The embedding size is the same as BERT hidden size. 
 \end{itemize}

Apart from the classification performance comparison of different kinds of annotator embeddings, we also conduct a detailed analysis of CTR-generated and history embeddings. Note that we do not use Learnt embeddings in this analysis since they are learnt end-to-end by the model during training. At first, we cluster each of the embedding types using K-means clustering. Afterwards, we compute the correlation between inter-annotator agreement (cohen's kappa) and cosine distance between every possible pair of annotator embeddings within and outside the clusters. Intuitively, an efficient annotator embedding should be similar for annotators with high inter-annotator agreement or, in other words, correlation should be strongly negative since cosine distance would be smaller between annotators with higher agreement.

\subsection{Model ablation: label text}

We also test the same models on different label texts. In addition to our experiments evaluating the impact of using label texts over not using them, the objective of testing different label texts is to assess the impact of using more or less relevant label texts. On both datasets, five conditions were tested:
\begin{enumerate}
    \item \textbf{``hate", ``not"}. The default setting on \textit{Dynabench}; inaccurate description condition on \textit{Guest}.
    \item \textbf{``misogynistic", ``not"}. The default setting on \textit{Guest}; inaccurate description condition on \textit{Dynabench}.
    \item \textbf{``hate", ``not hate"}. A variation of the first condition, to test whether a neutral label or an antonym label is better as the negative class for the domain-specific conditions.
    \item \textbf{``misogynistic", ``nonmisogynistic"}. A variation of the first condition, to test whether a neutral label or an antonym label is better as the negative class for the domain-specific conditions.
    \item \textbf{``yes", ``no"}. Non-domain-specific label text, to test the importance of domain-specificity in the label text for the model.
\end{enumerate}

In summary, the comparison between 1, 2, 3, 4 and 5 tests whether domain-specific labels are needed for the model; the comparison between 1, 3 and 2, 4 tests the effect of accurate domain specific class description; the comparison between 1, 2 and 3, 4 tests whether negative instances in different datasets can be represented in the same way.

\subsection{Training Settings}

We implement AnnoBERT and the BERT baseline using the Huggingface transformers library \cite{wolf2020transformers}. Models are trained for 4 epochs (except LEAM for 15 epochs), with a learning rate of $1e^{-5}$, cross-entropy loss, and a batch-size of 32. \textit{Adam} \cite{kingma2014adam} is used as the optimiser, with \textit{Tanh} activation function. LEAM model uses FastText embeddings of dimension 300 \citep{grave2018learning} as the encoder, and softmax function in the output layer. We report average results for 10 separate runs of each model.

\subsection{Evaluation Metrics}

Given the imbalanced nature of the hate speech classification task, we use the following metrics to evaluate our model and other baselines:

\textbf{Macro F1}: Macro F1 is the arithmetic mean of all the per-class F1 scores.

\textbf{Sensitivity} of the positive class: also known as true positive rate or recall of the positive class, is the probability of a positive test, given a positive truth. 

\textbf{Specificity} of the positive class: also known as true negative rate or the recall of the negative class, is the probability of a negative test, given a negative truth. 

We use macro F1 as the high-level representation of the overall model performance, but contrast sensitivity and specificity to reflect a model's bias towards either the positive or the negative class.

\section{Results and Discussions}

Basic model performances are summarised in Table \ref{tab:f1s}.

\begin{table}[htb]
\centering
\begin{tabular}{p{0.2\columnwidth}|p{0.2\columnwidth}p{0.2\columnwidth}p{0.2\columnwidth}}
\hline
Model      & Macro F1       & Sensitivity     & Specificity   \Tstrut\Bstrut   \\ 
\hline
\hline
\multicolumn{4}{c}{Dynabench}  \Tstrut\Bstrut \\ \hline
BERT       & \underline{76.04±0.17} & 79.43±0.82  & \underline{72.58±0.66}  \Tstrut\\
CrowdLayer & 42.77±1.87 & 68.52±10.14 & 31.65±10.12  \\
LEAM       & 39.37±0.94 & \textbf{94.64±1.42}  & 5.93±1.49    \\
AnnoBERT   & \textbf{76.19±0.20} & \underline{75.64±1.24}  & \textbf{77.17±1.16} \Bstrut  \\ \hline
\hline
\multicolumn{4}{c}{Guest}  \Tstrut\Bstrut \\ \hline
BERT       & \underline{66.26±1.77} & \underline{30.20±3.88}  & \textbf{97.53±0.41}   \Tstrut \\
CrowdLayer & 48.91±0.49 & 9.12±2.33   & 90.02±2.52   \\
LEAM       & 50.62±0.66 & 8.82±2.72   & 94.40±1.91   \\
AnnoBERT   & \textbf{69.31±0.57} & \textbf{37.26±1.90}  & \underline{96.81±0.42} \Bstrut \\
\hline
\end{tabular}
\caption{Performance of AnnoBERT model and baselines on two datasets. Macro F1, sensitivity, and specificity scores, mean from 10 runs each, with standard errors on mean. Best performance \textbf{bolded}, second best performance \underline{underlined}. }
\label{tab:f1s}
\end{table}

\textbf{The main weakness shared by baselines is a bias towards the majority class, mediated by class imbalance severity.}

All baselines have higher sensitivity than specificity on \textit{Dynabench} and higher specificity than sensitivity on \textit{Guest}. 
On the much more imbalanced \textit{Guest} (Table \ref{tab:datasets}), a much lower sensitivity is observed in all baselines. This contributed greatly to overall lower macro F1 scores.
BERT is the most robust baseline on this front, yet still achieving only 30.20 sensitivity  on \textit{Guest}and 72.58 specificity on \textit{Dynabench} respectively. LEAM and CrowdLayer are consistently worse than BERT. 
These highlight the challenge of class imbalance in hate speech detection, and that transformer models are strong baselines even compared to alternative models with additional annotator and label features.

\textbf{AnnoBERT improves performance through addressing the challenge of class imbalance.} 

As shown in Table \ref{tab:f1s}, on both datasets, AnnoBERT achieves the highest recall of the minority class -- specificity on \textit{Dynabench} and sensitivity on \textit{Guest}, showing better robustness against class imbalance. Although with a small trade-off on the complementing metric, the advantage with the minority class is reflected on the overall performance as a higher macro F1 on both datasets, with the advantage much larger on the highly imbalanced, realistically sampled \textit{Guest} than the balanced, synthetically constructed \textit{Dynabench}. 

On the imbalanced subset of \textit{Dynabench} (Table \ref{tab:imba}), model relative performances follow the same pattern as those on \textit{Guest}, with AnnoBERT having considerable advantage on sensitivity and in turn macro F1. This clarifies that the performance difference between \textit{Guest} and \textit{Dynabench} is indeed mainly caused by different class distributions. 

These results altogether show that AnnoBERT is less biased against the majority class and benefits hate speech detection the most when the data is highly imbalanced. Specifically, it is better at picking up the characteristics of the minority class through limited instances during training, resulting in higher recall of that class.

\begin{table}[htb]
\centering
\begin{tabular}{p{0.2\columnwidth}|p{0.2\columnwidth}p{0.2\columnwidth}p{0.2\columnwidth}}
\hline
Model    & Macro F1        & Sensitivity     & Specificity    \Tstrut\Bstrut  \\ 
\hline
BERT     & 54.17±1.08 & 9.12±1.78  & \textbf{98.10±0.44} \Tstrut \\
AnnoBERT & \textbf{59.69±0.65} & \textbf{25.00±2.22} & 93.93±1.00  \Bstrut \\
\hline
\end{tabular}
\caption{Performance of AnnoBERT and BERT on imbalanced sample of \textit{Dynabench}. Macro F1, sensitivity, and specificity scores, mean from 10 runs each, with standard errors on mean. Better performance \textbf{bolded}.}
\label{tab:imba}
\end{table}

\textbf{AnnoBERT has the most advantage on edge cases which annotators disagree on. }

The comparison between AnnoBERT and BERT on all instances and only instances with disagreement in \textit{Guest} is shown in Table \ref{tab:disagreement}.

\begin{table}[htb]
\centering
\begin{tabular}{p{0.2\columnwidth}|p{0.2\columnwidth}p{0.2\columnwidth}p{0.2\columnwidth}}
\hline
Model                                           & Macro F1   & Sensitivity & Specificity \Tstrut\Bstrut \\ 
\hline
\hline
\multicolumn{4}{c}{Guest, disagreement}     \Tstrut\Bstrut    \\ 
\hline
BERT                                            & 53.33±2.71 & 30.20±3.88  & \textbf{93.97±1.43}  \Tstrut \\
AnnoBERT                                        & \textbf{58.44±1.18} & \textbf{37.26±1.90}  & 92.79±1.21  \Bstrut \\ 
\hline
\hline
\multicolumn{4}{c}{Guest, overall}      \Tstrut\Bstrut     \\ 
\hline
BERT                                            & 66.26±1.77 & 30.20±3.88  & \textbf{97.53±0.41}  \Tstrut \\
AnnoBERT                                        & \textbf{69.31±0.57} & \textbf{37.26±1.90}  & 96.81±0.42  \Bstrut \\
\hline
\end{tabular}
\caption{Performance of AnnoBERT and BERT on instances with disagreement and all instances from \textit{Guest}. Macro F1, sensitivity, and specificity scores, mean from 10 runs each, with standard errors on mean. Better performance \textbf{bolded}.}
\label{tab:disagreement}
\end{table}

Both models had lower specificity on the subset with disagreement, which is within expectation as these are instances that also confused human annotators. 

Model relative performance is similar on disagreement in a sense that AnnoBERT marginally trades off specificity for a large improvement on sensitivity and macro F1. 
However, the advantage of AnnoBERT over BERT is noticeably larger on the disagreement instances (+5.11 macro F1) than on the whole dataset (+3.15 macro F1). This highlights the remarkable benefit of using AnnoBERT's annotator and label information to improve hate speech detection in edge cases.

\subsection{Model Ablation}

\textbf{Theoretically grounded CTR embeddings benefit the classification task more than end-to-end trained embeddings and raw annotation history.}

Models with different annotator embeddings are compared in Table \ref{tab:annotator_embedings}.

\begin{table}[htb]
\centering
\begin{tabular}{p{0.2\columnwidth}|p{0.2\columnwidth}p{0.2\columnwidth}p{0.2\columnwidth}}
\hline
Annotator embedding & Macro F1   & Sensitivity & Specificity \Tstrut\Bstrut \\ 
\hline
\multicolumn{1}{l}{Dynabench} &            &             &            \Tstrut\Bstrut  \\ 
\hline
CTR, uf             & \underline{76.05±0.24} & \underline{73.46±1.31}  & \textbf{79.40±1.19}  \Tstrut \\
CTR, f              & \textbf{76.19±0.20} & \textbf{75.64±1.24}  & 77.17±1.16   \\
History, uf         & 38.52±2.60 & 41.91±13.69 & 58.81±13.63  \\
History, f          & 39.22±2.76 & 22.60±9.55  & \underline{78.24±8.93}   \\
Learnt, uf          & 64.83±1.03 & 55.60±2.58  & 76.57±1.87  \Bstrut \\ 
\hline
\multicolumn{1}{l}{Guest} &             &             &         \Tstrut\Bstrut     \\ 
\hline
CTR, uf             & \underline{69.05±0.58} & \textbf{37.45±2.10}  & 96.62±0.37  \Tstrut \\
CTR, f              & \textbf{69.31±0.57} & \underline{37.26±1.90}  & \textbf{96.81±0.42}   \\
History, uf         & 65.41±0.70 & 35.59±2.38  & 94.52±1.06   \\
History, f          & 65.73±0.89 & 30.49±2.60  & \underline{96.78±0.58}   \\
Learnt, uf          & 66.36±0.52 & 31.86±2.57  & 96.69±0.69 \Bstrut \\
\hline
\end{tabular}
\caption{Performance of AnnoBERT model variants with different annotator embeddings. Macro F1, sensitivity, and specificity scores, mean from 10 runs each, with standard errors on mean. f/uf: annotator embeddings frozen/unfrozen during task-specific training. Best performance \textbf{bolded}, second best performance \underline{underlined}. }
\label{tab:annotator_embedings}
\end{table}

Overall, the original AnnoBERT model with CTR embeddings consistently outperforms the ablation models with alternative embeddings, with the end-to-end learnt embeddings better than using annotation history. 

Annotator embeddings trained end-to-end are not affected by annotator frequencies, but is inferior to a theoretically driven model in the overall medium classification performance. 
Only the CTR-based models consistently outperform the BERT baselines, meaning that with a strong baseline, additional annotator information is only beneficial with high quality annotator embeddings.

Frozen annotator embeddings tend to achieve higher mean macro F1 scores, but the difference is marginal considering the standard errors. Thus, the model does not forget much about the annotators during end-to-end training of the model.

\textbf{CTR embeddings reflect annotator agreement better than embeddings generated with annotator history.}

\begin{figure*}[ht!]
\centering
    \begin{subfigure}{0.34\textwidth}
      \centering
      \includegraphics[width=\linewidth]{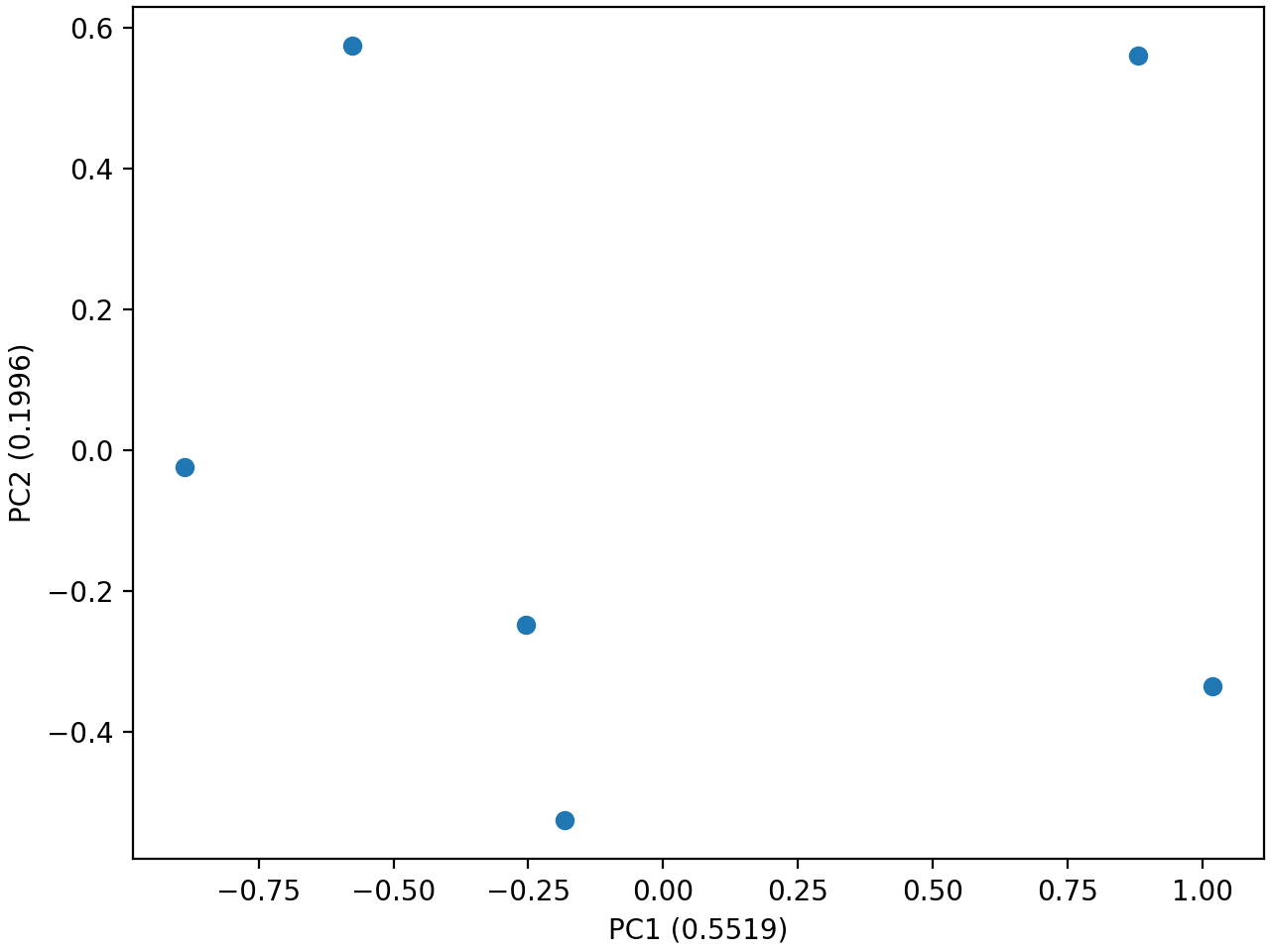}
      \caption{Annotator embeddings using CTR model.}
      \label{fig:ctr-guest}
    \end{subfigure}
    \begin{subfigure}{0.34\textwidth}
      \centering
      \includegraphics[width=\linewidth]{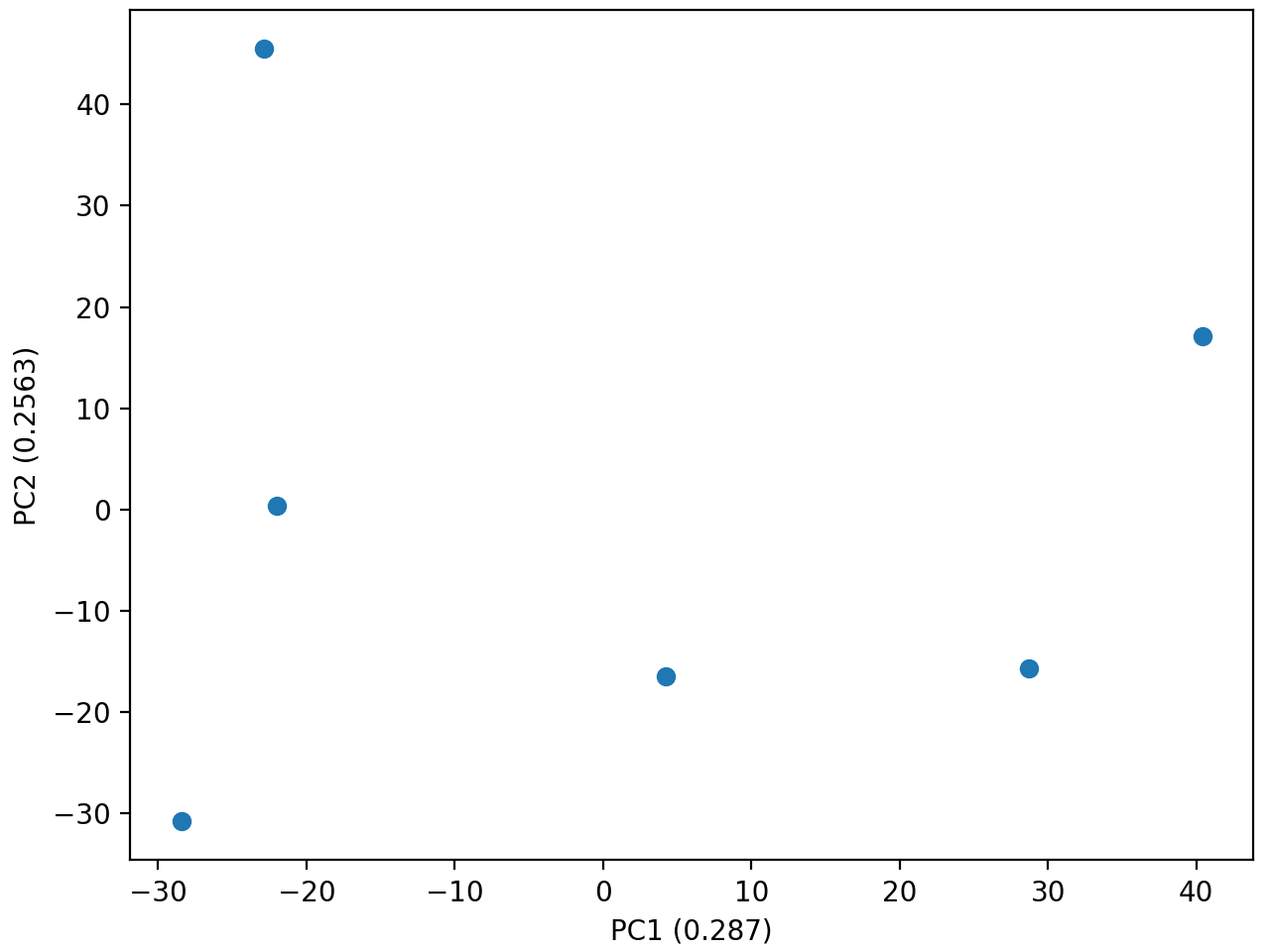}
      \caption{Annotator embeddings using annotators' history.}
      \label{fig:history-guest}
    \end{subfigure}
\caption{Visualisation through PCA dimension reduction of \textit{Guest} dataset. Variance explained by each principal component is shown within brackets.}
\label{fig:annotators-clusters-guest}
\end{figure*}

\begin{figure*}[ht!]
\centering
    \begin{subfigure}{0.34\textwidth}
      \centering
      \includegraphics[width=\linewidth]{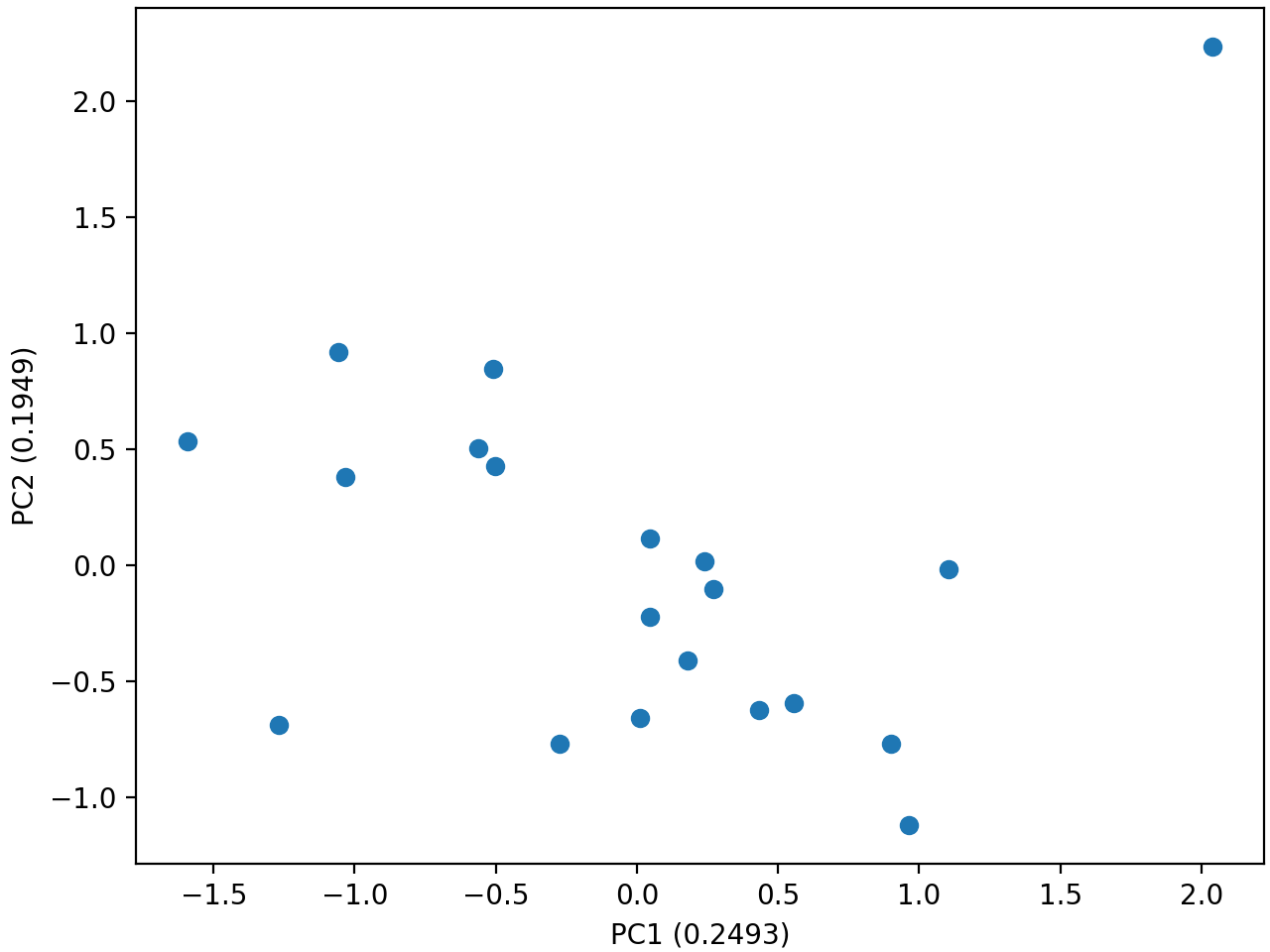}
      \caption{Annotator embeddings using CTR model.}
      \label{fig:ctr-dynabench}
    \end{subfigure}
    \begin{subfigure}{0.34\textwidth}
      \centering
      \includegraphics[width=\linewidth]{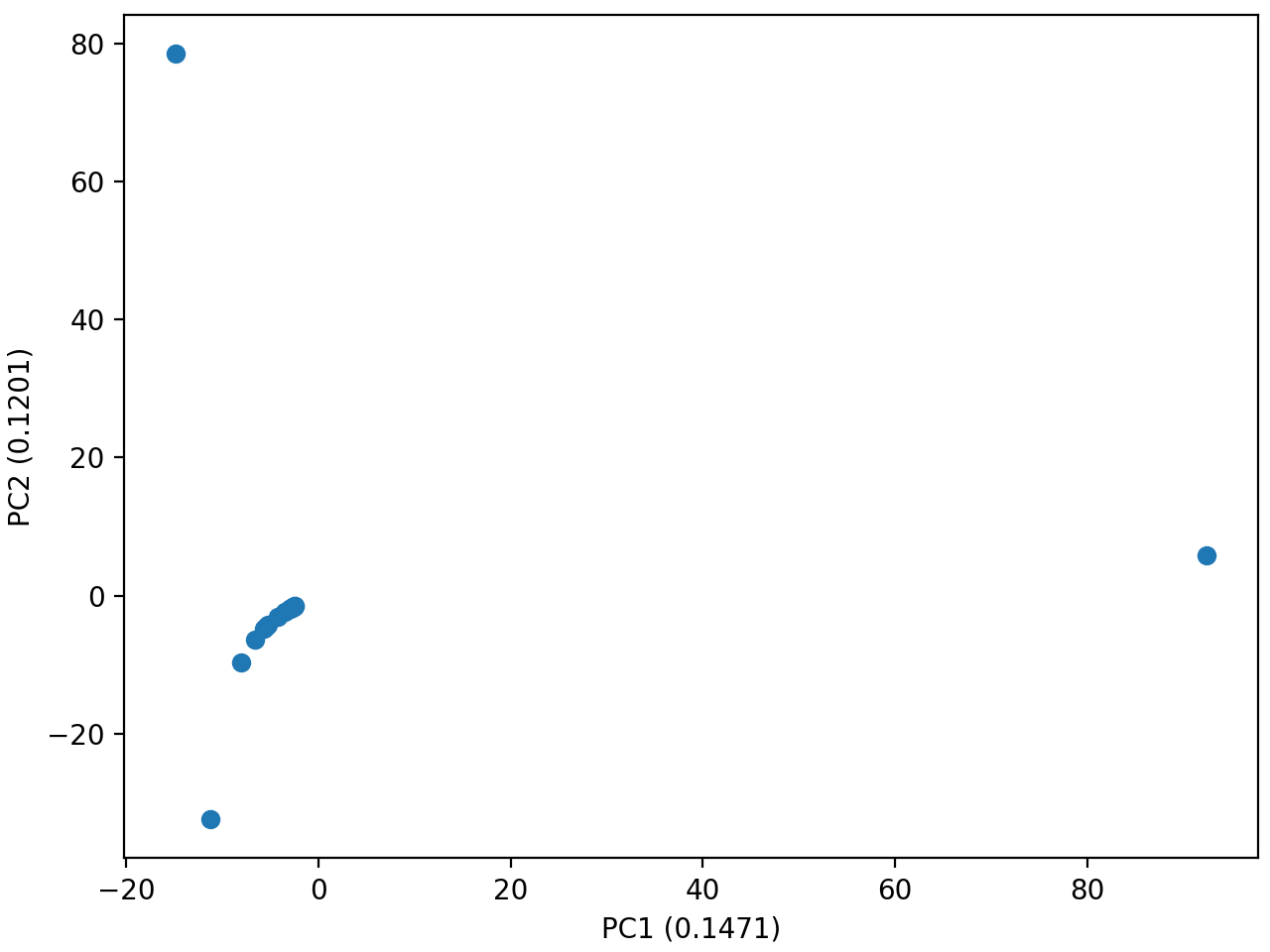}
      \caption{Annotator embeddings using annotators' history.}
      \label{fig:history-dynabench}
    \end{subfigure}
\caption{Visualisation through PCA dimension reduction of annotator embeddings on \textit{Dynabench} dataset. Variance explained by each principle component is shown within brackets.}
\label{fig:annotators-clusters-dynabench}
\end{figure*}

Figures \ref{fig:annotators-clusters-guest} and \ref{fig:annotators-clusters-dynabench} visualise the annotator embeddings, dimension-reduced through Principal Component Analysis. For both datasets, variance explained by the first two principal components (in brackets) is higher for annotator embeddings using the CTR model than using the annotators' history, as shown in Figures~\ref{fig:ctr-guest} and \ref{fig:ctr-dynabench}.
Comparing \textit{Dynabench} to \textit{Guest}, there are a lot more unique annotators (20 as opposed to the Guest dataset which only has 6 annotators), and each instance is only associated with one annotator. As a result, the relative locations of the annotators in the embedding space computed through raw history, as shown in Figure~\ref{fig:history-dynabench}, is largely determined by the volume of annotations each annotator made. For example, 3 annotators annotated a large portion (almost $35\%$) of the instances in Dynabench and hence these annotators are separated far apart in the embedding space. All the remaining 17 annotators, who produced fewer annotations, are close together in the embedding space because they have lots of missing annotations (0s) in the raw history embeddings. In comparison, the CTR embeddings are robust to this factor and do not just reflect the volume of annotations or the missing annotations (0s), but also the annotators' label choices via topic modelling. This likely underlies the classification difference across different annotator embeddings, which is much larger in \textit{Dynabench} than \textit{Guest}.

Table~\ref{tab:emb_ablation} shows the correlation of inter-annotator agreement (cohen's kappa) with cosine distance between all the possible pairs of annotators. CTR embeddings better reflect agreement since their correlation is strongly negative as compared to history embeddings. Correlation is expected to be negative since annotators with higher agreement are more similar and hence, their cosine distance will be smaller.

\begin{table}[htb]
\centering
\begin{tabular}{l|c}
\hline
Embedding type      &  Pearson's r  \Tstrut\Bstrut \\    
\hline
  History &  -0.2956   \Tstrut \\  
  CTR &  -0.4031  \Bstrut \\    
\hline
\end{tabular}
\caption{Correlation, measured by Pearson's r, of inter-annotator agreement with cosine distance between possible pairs of annotators for \textit{Guest} dataset.}
\label{tab:emb_ablation}
\end{table}

\textbf{The exact label text contributes less to the model performance than associating annotator embeddings with the label choices.}

Performance with alternative label texts are shown in Table \ref{tab:labels}.

\begin{table}[htb]
\centering
\begin{tabular}{p{0.2\columnwidth}|p{0.2\columnwidth}p{0.2\columnwidth}p{0.2\columnwidth}}
\hline
Label text                    & Macro F1   & Sensitivity & Specificity  \Tstrut\Bstrut \\ 
\hline
\multicolumn{1}{l}{Dynabench} &            &             &          \Tstrut\Bstrut    \\ 
\hline
hate, not hate                & 76.05±0.26 & 74.25±1.69  & \textbf{78.55±1.75}  \Tstrut \\
hate, not                     & \underline{76.19±0.20} & \underline{75.64±1.24}  & 77.17±1.16   \\
misogynistic, nonmisogynistic & 75.29±0.32 & 75.38±2.12  & 75.68±2.05   \\
misogynistic, not             & 76.03±0.24 & 74.52±1.42  & \underline{78.13±1.23}   \\
yes, no                       & \textbf{76.36±0.21} & \textbf{76.84±1.55}  & 76.24±1.71  \Bstrut \\ 
\hline
\multicolumn{1}{l}{Guest}     &            &             &          \Tstrut\Bstrut    \\ 
\hline
hate, not hate                & 68.97±0.64 & \underline{41.08±2.03}  & 95.44±0.56  \Tstrut \\
hate, not                     & 68.38±0.41 & 35.39±2.35  & \textbf{96.83±0.55}   \\
misogynistic, nonmisogynistic & 67.58±0.49 & 38.23±1.45  & 89.11±6.00   \\
misogynistic, not             & \underline{69.31±0.57} & 37.26±1.90  & \underline{96.81±0.42}   \\
yes, no                       & \textbf{70.17±0.52} & \textbf{44.41±2.77}  & 95.33±0.77  \Bstrut \\
\hline
\end{tabular}
\caption{Performance of AnnoBERT model and baselines on two datasets. Macro F1, sensitivity, and specificity scores, mean from 10 runs each, with standard errors on mean. Best performance \textbf{bolded}, second best performance \underline{underlined}. }
\label{tab:labels}
\end{table}

Label text has a very different effect than annotator embeddings. 
For one thing, classification performance does not vary as much -- the gap between the best and worst models is less than 3\% in macro F1 across label texts, in contrast to over 30\% between annotator embeddings. 
For another, the more varying dataset is \textit{Guest} here as opposed to \textit{Dynabench}. Our interpretation is that a dataset with a narrow domain (\textit{Guest}, which specialises in misogyny) is more sensitive to label texts. 

Surprisingly, in both datasets, ``yes" and ``no" texts appear to be the most effective, albeit the difference being marginal. The reason behind this could be that this pair is the most semantically contrasting, for both language models and audience from different backgrounds. ``Misogynistic" vs ``nonmisogynistic" is the worst performing.
However, pairing ``misogynistic"  with a neutral negative class text (``not") noticeably improves the classification performance, leading to a second highest mean macro F1 on \textit{Guest}. 
This is likely because the BERT embeddings with ``nonmisogynistic", a much rarer word than any other label texts, is of lower quality, as rare words are a major problem in contexualised embeddings \citep{schick2020rare}. 
Nonetheless, all label text variants outperform the baselines.

Overall, the two model ablation studies show that the advantage of AnnoBERT lies within the ability to  pair effectively encoded annotator attributes with contrasting classes.
The exact label text does not matter as much as annotator embedding quality, but it is likely to be beneficial to have text combinations that are less domain-specific, semantically contrasting, and have higher word frequency for the base transformer model. 

\textbf{Implications on the role of annotators in hate speech annotation and classification.}
It was not until recently that the annotators' role were brought into attention in abusive language detection \citep{basile_its_2021} and contrasting paradigms for subjective annotation were defined \citep{rottger_two_2021}\footnote{These definitions were not available when we carried out the experiments.}: while a descriptive paradigm aims to reflect individuals' viewpoints by providing less instruction, a prescriptive paradigm aims to reflect one predefined taxonomy by providing extensive instruction. Both \textit{Dynabench} and \textit{Guest} datasets follow a prescriptive paradigm with clear class definitions. 

On the one hand, our experiments highlight the significant role that annotators play in the creation of subjective data. More precisely, we showed that even with prescriptively constructed datasets, where annotators are expected to disagree much less, annotator information can considerably benefit classification performance, especially on edge cases where disagreement inevitably happens. This calls for the field to include and analyse annotator behaviour as well as to utilise such information when approaching classification tasks, particularly tasks of a subjective nature. Admittedly, privacy is an important concern when it comes to publicly releasing annotator information, which is why our study focuses on using only annotators' judgements rather than their personal information in the classification model. 

On the other hand, our study also points towards a promising direction for identifying and mitigating annotator biases. One example use case would be that, with annotator representations that better represent the level of agreement between them, annotators who perform differently from the others can be more easily identified through visualisation. 
\section{Conclusions and Future Work}

Our work is the first to integrate annotator characteristics, text and label representations in one hate speech detection model.
Our proposed model, AnnoBERT, improves performance in hate speech detection on top of a transformer-based baseline. AnnoBERT works especially well on imbalanced data, which is characteristic of nearly all extant hate speech datasets. The scarcer the data of the minority class, the larger its advantage on the recall of the minority class and the more visible the improvement in overall classification metric. We clarified this effect of class distribution using an imbalanced sample of an originally balanced dataset. Through further analysis, we show that our proposed approach benefits especially instances on which annotators disagree more. Through ablation studies, we showed the relative contributions of annotator embeddings and label text to the model performance, and tested a range of alternative annotator embeddings and label text combinations. 

Although we used BERT as the base model for our proposed AnnoBERT, we expect the benefit of our approach to translate to other transformer-based models; although we used CTR user latent vectors as annotator embeddings, this approach shows potential with other user models. Thus, the most immediate future steps would be using more advanced transformer models and user models, especially user models that directly link text features to user ratings, which will produce inherently explainable hate speech detection models.

Additionally, our approach carries significant practical value, with its considerable advantage on learning from limited minority class samples -- in the real world, abusive content amounts to around 0.1\% to 3\% of social media \citep{founta2018large}. Our results encourage future hate speech detection research to consider annotator preference and label information to address this common challenge. More importantly, our approach enables customised hate speech detection on an aggregated level. So far, hate speech detection models used separate classifiers to work on different standards for hate speech. In contrast, our approach enables prediction of different groups of annotators through a single model by only replacing the annotator embeddings without the need of replacing the main detection model. Likewise, our approach does not require collection of annotator demographics as it solely relies on their annotation histories.

Possible future directions include utilising user models that directly link text features to user ratings, which will produce explainable hate speech detection models, or scaling up our approach to, for example, predicting moderation choices of communities with different moderation strategies.

\section{Ethics Statement}

Hate speech online is a sensitive issue, and there are some ethical issues in the controversy on free speech. To further improve the fairness and reliability of our work, we claim the following ethical considerations:

\begin{itemize}
    \item \textbf{Confidentiality:} Access to data is critical to the effectiveness of our work. Since the data is already publicly available, we have replaced all personal data with a special token $<$user$>$ to ensure user anonymity. The information of all the annotators corresponding to the dataset is hidden, and only the ID number of the annotator and its annotation content are displayed.
    \item \textbf{Potential for harm:} We do not intend to harm vulnerable groups that are already discriminated against based on specific characteristics. Our work serves only the benign purpose of detecting and mitigating abusive expressions.
    \item \textbf{Results communication:} We guarantee that there is no plagiarism or academic misconduct, but we acknowledge that there may be limitations and potential misrepresentations when analysing social media data, particularly for those abusive data that does not clearly represent the target of the attack.
\end{itemize}

It can be debatable whether all annotator perspectives should be treated equally, especially when they lack knowledge or relevance to the speech at question. If annotators are random samples from the population, it is likely that they diverge from the opinions of the minority groups targeted by the online hate speech. In an extreme scenario, for example, treating a white supremacist annotator and an oppressed individual with the same weights can further suppress the voice of the latter group. Future work from a social science perspective is needed.

Moreover, adversaries, who actually spread hate speech online, can use this kind of research for malicious purposes such as to learn how to prevent being detected.

\bibliography{custom}





\end{document}